\pdfoutput=1

\documentclass[11pt]{article}

\usepackage{acl}
\usepackage{caption}
\usepackage{times}
\usepackage{latexsym}
\usepackage{float}
\usepackage[T1]{fontenc}

\usepackage[utf8]{inputenc}

\usepackage{microtype}

\usepackage{inconsolata}

\usepackage{graphicx}
\usepackage{tikz}
\usepackage{float}
\title{Semantic, Orthographic, and Phonological Biases in Humans' Wordle Gameplay}

\author{Jiadong Liang, Adam Kabbara, Jiaying Liu, Ronaldo Luo, Kina Kim,  Michael Guerzhoy \\
         Division of Engineering Science, University of Toronto\\Toronto, Ontario, Canada\\
            \texttt{esgary.liang@mail.utoronto.ca, adam.kabbara@mail.utoronto.ca,}\\ \texttt{cindyjy.liu@mail.utoronto.ca, ronaldo.luo@mail.utoronto.ca,}\\ \texttt{kina19131@gmail.com, guerzhoy@cs.toronto.edu}}

\begin{document}
\maketitle
\begin{abstract}
    We show that human players' gameplay in the game of Wordle is influenced by the semantics, orthography, and phonology of the player's previous guesses. We compare actual human players' guesses with near-optimal guesses using NLP techniques. We study human language use in the constrained environment of Wordle, which is situated between natural language use and the artificial word association task.
\end{abstract}

\section{Introduction}

%

Wordle is a daily word-guessing game where players attempt to identify an unknown five-letter word within six attempts \cite{wordle2021}. Players usually attempt to minimize the number of guesses they make. Players also usually want to maintain a ``streak" of having solved the game within at most 6 guesses for several days. 


We explore the difference between near-optimal play and human gameplay, which may be influenced by cognitive shortcuts and biases. In order to estimate near-optimal plays, we use the maximum-entropy heuristic. We verify that the heuristic is near-optimal.

In settings where word association is important, humans are influenced by salient past information, a phenomenon known as \textit{priming} in psychology~\cite{schacter1998priming}. We conjecture that priming effects exist in the game of Wordle as well. Additionally, we conjecture that humans will tend to depart less from previous guesses in order to minimize cognitive load.

We review the prior work on priming in psychology, and in particular, how priming influences future word choice. We then review the optimal strategy in Wordle, as well as heuristics that approximate it. Following this discussion, we introduce our human guess data. We then present our approach to measuring human biases in Wordle gameplay and demonstrate the systematic differences between human play are near-optimal play.




\section{Background: Human Cognitive Processes \label{priming}}
Priming is a phenomenon in psychology where past experience influences behavior without the person's explicit knowledge of the influence~\cite{schacter1998priming}. Priming sometimes manifests in word association. Prior works have demonstrated the grammatical class, semantic meaning and rhyme of the previous (cue) word would influence the later (response) word by humans.
This effect has also been explained through models of lexical access in speech production, where residual activation from previously retrieved lexical items can facilitate or bias subsequent retrievals~\cite{levelt1989speaking, roelofs1992spreading}. 

\citet{deese1962form} and \citet{de2008word} explore the effect of the grammatical class of cue words over the response word.


\citet{steyvers2005large} use data collected by~\citet{nelson1999university} to demonstrate that the word association network --- a graph where two words are connected if they are members of a cue-response pair --- is sparse, with each word connected to only 0.44\% of other words. They also use data collected by \citet{miller1995wordnet} and \citet{fellbaum1998wordnet}, and found that the word network constructed based on semantics of words exhibits sparseness, connectedness, neighboring clustering, and power-law degree distribution. All of these are the same characteristics exhibited in the free association network.




\citet{nelson1987prior} demonstrate the effect of rhyme on memory and word association. They ran an experiment where subjects would initially study (read aloud) the cue-target pair of a given rhyme; then 1.5-2 minutes after they finished studying, a meaning-related cue word and its semantic relation with the target word would be given and the participants would be required to read it aloud and recall the word they studied. In the experiment, cue words that rhyme with many other words would decrease the accuracy of the respondent, regardless of the meaning-related cue word. Through conducting a further experiment that changed all the cue-target pairs studied to be meaning-related and only half to be also rhyme-related, \citet{nelson1987prior}  showed that the effect of rhyming appears only if the subjects actively attend to it when studying the word pairs.

\citet{matusevych2018analyzing} studied human word association based on word attributes. \citet{luo2025automatically} demonstrated it is possible to weakly predict human amusement responses to Wordle gameplay using features similar to the ones used in this study.

Related work on word puzzle solving provides further evidence that lexical retrieval and phonological awareness are central to constrained word-generation tasks. \citet{underwood1994expert} found that expert crossword solvers outperform intermediate solvers in generating candidate words, solving anagrams, and manipulating morphemic and syllabic components of words. These skills are closely tied to efficient lexical access and morphological processing. These findings suggest that in tasks such as Wordle, players may similarly rely on lexical retrieval strategies and phonological structure to guide their guesses.

\section{Background: Wordle solving mechanisms \label{wordle}}

The objective of  Wordle depends on the player --- it can be maintaining the streak (i.e., trying not to lose today's game), winning in as few guesses as possible, or even winning the game using funny words. 

However, most of the solving mechanisms are designed to optimize objectives regarding the number of guesses, such as minimizing the average number of guesses, minimizing the number of guesses in the worst case, etc. Those mechanisms can be classified into two classes: the exact optimization approach and heuristic approaches. The best approaches based on heuristics achieve results that are only marginally inferior to exact methods.


\citet{bertsimas2024exact} found an optimal and efficient solution for Wordle that minimizes the average number of guesses using dynamic programming. They show that the word \verb;"SALET"; is the best starting guess and the minimum average number of guesses required is 3.421. They demonstrate that their optimal approach never results in a loss (i.e., the algorithm always completes the game within 6 guesses).

\textbf{Heuristic approaches} to Wordle can achieve performance that is very close to optimal. The depth-1 \texttt{minimax} heuristic aims to minimize the number of guesses for the worst case, with search depth of 1. For each guess, it iterates through all possible words in the game and chooses the one that minimizes the size of maximum partition (the amount of possible solutions after the current guess) as the guess. Given the starting guess as \verb;"SALET";, it is guaranteed to finish the game in 5 guesses, and has the average number of guesses of 3.482 \cite{CatchemAL_2022}.  The entropy-based heuristic (also with depth 1) reduces the uncertainty at each step by choosing the guess that decreases (on average) the most number of potential solutions after that guess \cite{shannon1948mathematical} \cite{CatchemAL_2022}. It is also guaranteed to complete the game in 6 guesses and have the average number of guesses of 3.432. We use the Doddle\footnote{\url{https://github.com/CatchemAL/Doddle}} implementation to report those results.




\section{Data \label{data}}
The human guess data was sourced from Reddit. The machine-generated guesses are obtained using Doddle, an open-source Wordle solver introduced earlier. Although an ideal comparison would be with the optimal model, the Doddle solver was chosen for computational reasons. It's important to note that the performance difference between the exact dynamic programming solution and the heuristic entropy solver is minimal: the exact solution achieves a minimum average of 3.421 guesses, while the heuristic-based solver has an average of 3.482 guesses for its minimax heuristic and 3.432 guesses for its entropy-based heuristic. Doddle's heuristic min-entropy solver we use will be referred to as the \textit{near-optimal} strategy.

\subsection{Data collection}
The data is collected from the \texttt{r/Wordle} subreddit, where people share their guesses online, contributing to a total of 83,000 data entries ~\cite{reddit_data_source}. We extract games posted by people in the commonly-used format on the subreddit using a regular expression.





\section{Methods}

\subsection{Measuring Human Biases} 
To quantitatively assess the influence of human cognitive biases in Wordle games, human plays are compared to their entropy-based near-optimal counterpart, where five different metrics described below are used to reveal different aspects of human biases (semantic, orthographic, and phonological). For each guess in the data, the metrics below are computed through comparing that guess with the previous one (instead of comparing with all prior guesses) unless otherwise stated. 



\subsubsection{Levenshtein Distance}
The Levenshtein Distance measures the minimum number of edits --- insertions, deletions, or substitutions --- needed to transform one word into another ~\cite{levenshtein1966binary}. This feature captures how closely a player's subsequent guesses align with their previous ones in terms of structural similarity. A smaller Levenshtein distance may indicate that the player is selecting guesses that are more similar to their prior attempts, potentially reflecting a reluctance to explore novel letter combinations or a preference for minimizing cognitive effort.

\subsubsection{Semantic Distance}
The GloVe~\cite{pennington2014glove} distance is computed using the negative cosine similarity between GloVe word embedding pairs. Humans may potentially be biased to make guesses that are semantically close to their previous guesses.


\subsubsection{Hamming Distance} 
The Hamming distance between words is the number of locations where the words differ. We conjecture that guesses that are close to previous guesses in the Hamming distance sense are easier to make.


\subsubsection{Rhyme}

To determine whether two words rhyme or not, their phonic transcription was used. This was achieved using the \verb;pronouncing; library, which provides a phonetic transcription based on the CMU Pronouncing Dictionary \cite{CMU}. Two words are considered to have a \textit{perfect rhyme} if they have matching phonetic endings which include stressed vowels \cite{PerfectRhyme}. We assess whether the guess rhymes with the previous one.


\section{Experiments}


We compare how human guesses/plays differ systematically from near-optimal plays. We obtain distributions of human plays and near-optimal plays, and compare them. We assess the effect size (difference between the distributions) using Cohen's d, and we computed the p-values based on the t-statistics for the difference between the two distributions.

We analyze separately games starting from different positions. We use the notation $c_g g c_y y c_b b$ and $(c_g, c_y, c_b)$, where the number of ``green" guesses (correct letter in the correct place) is denoted with $c_g$, the number of ``yellow" guesses (correct letter in the incorrect place) is denoted with $c_y$, and the number of letter guesses that are incorrect is $c_b$.

\begin{figure*}[!ht]
    \centering
    \begin{tabular}{@{}cc@{}}
        \includegraphics[width=0.45\textwidth]{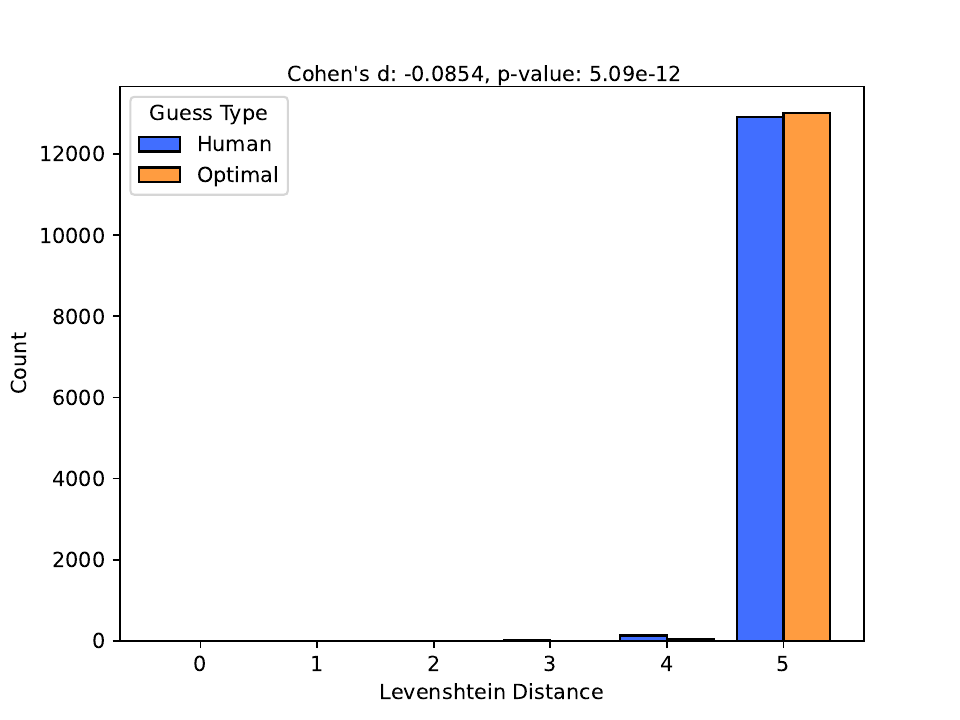} &
        \includegraphics[width=0.45\textwidth]{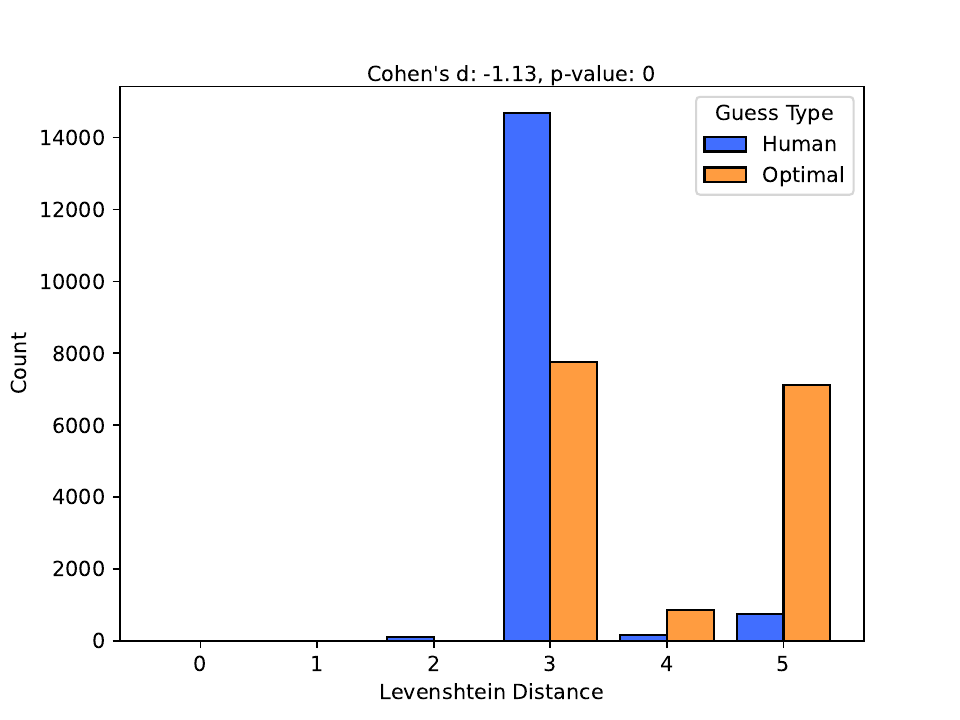} \\
        \parbox{0.45\textwidth}{\small\centering (a) Levenshtein distance between human guesses and near-optimal guesses for 0g0y5b: both choose distance 5 most of the time. Humans suboptimally play letters they know aren't there.} &
        \parbox{0.45\textwidth}{\small\centering (b) Levenshtein distance between human guesses and near-optimal guesses for 2g0y3b: humans underexplore compared to near-optimal.} \\[1cm]
        
        \includegraphics[width=0.45\textwidth]{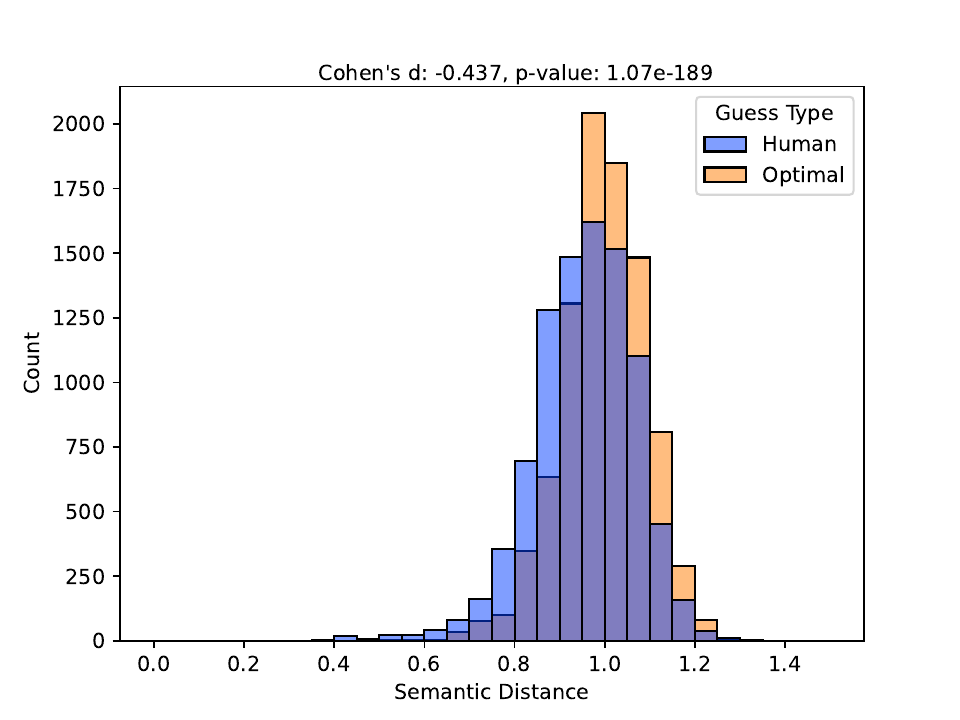} &
        \includegraphics[width=0.45\textwidth]{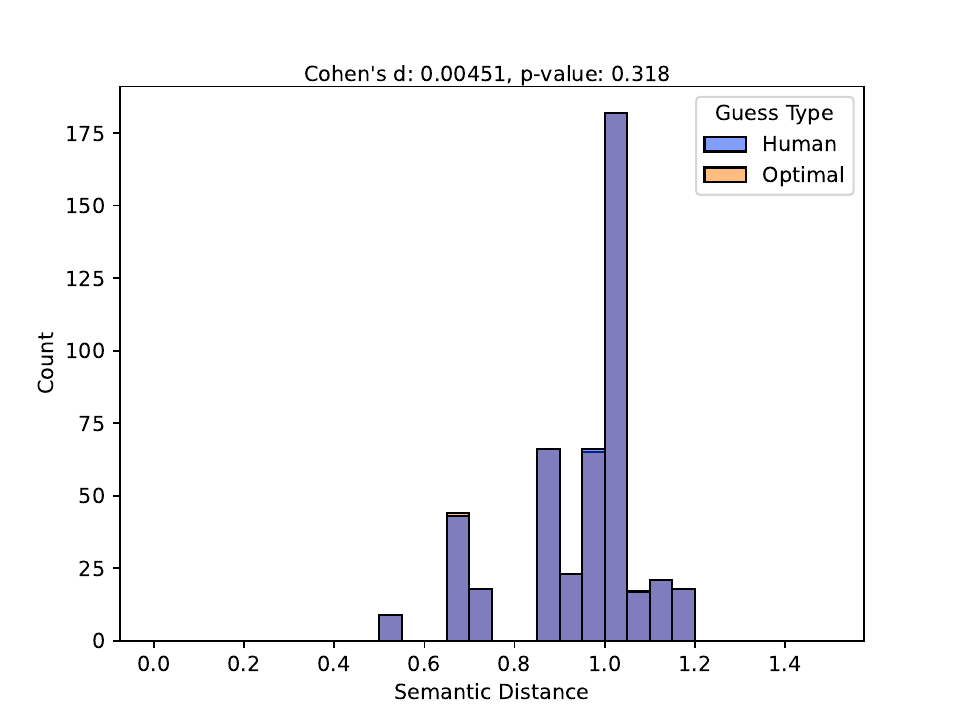} \\
        \parbox{0.45\textwidth}{\small\centering (c) Semantic distance between human guesses and near-optimal guesses for 0g0y5b: humans slightly biased towards underexploring} &
        \parbox{0.45\textwidth}{\small\centering (d) Semantic distance between human guesses and near-optimal guesses for 3g2y0b: no bias.} \\[1cm]
        
        \includegraphics[width=0.45\textwidth]{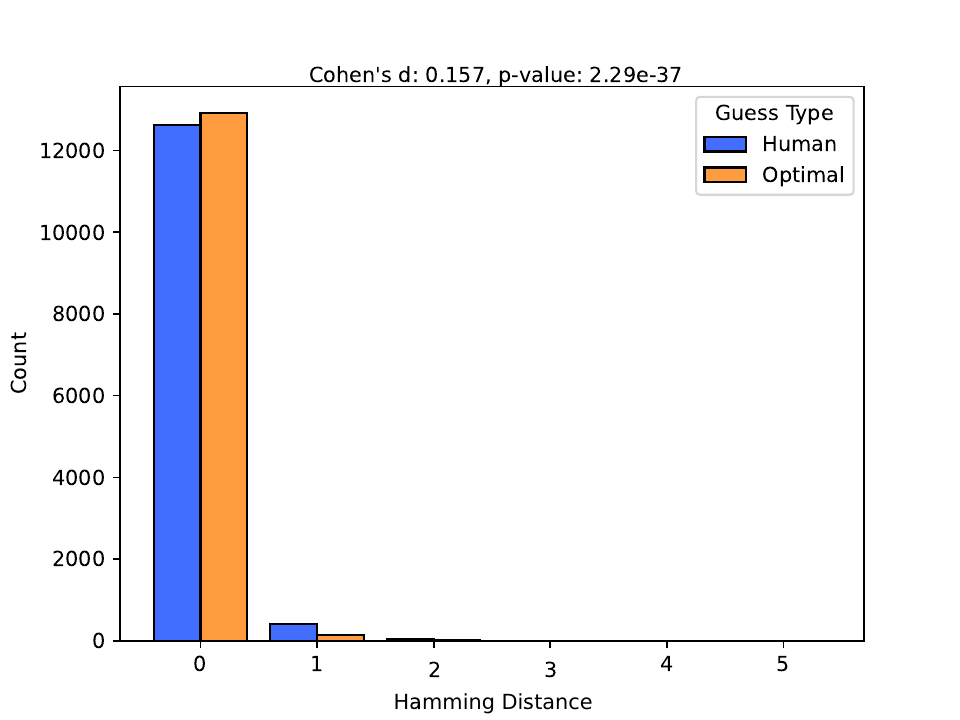} &
        \includegraphics[width=0.45\textwidth]{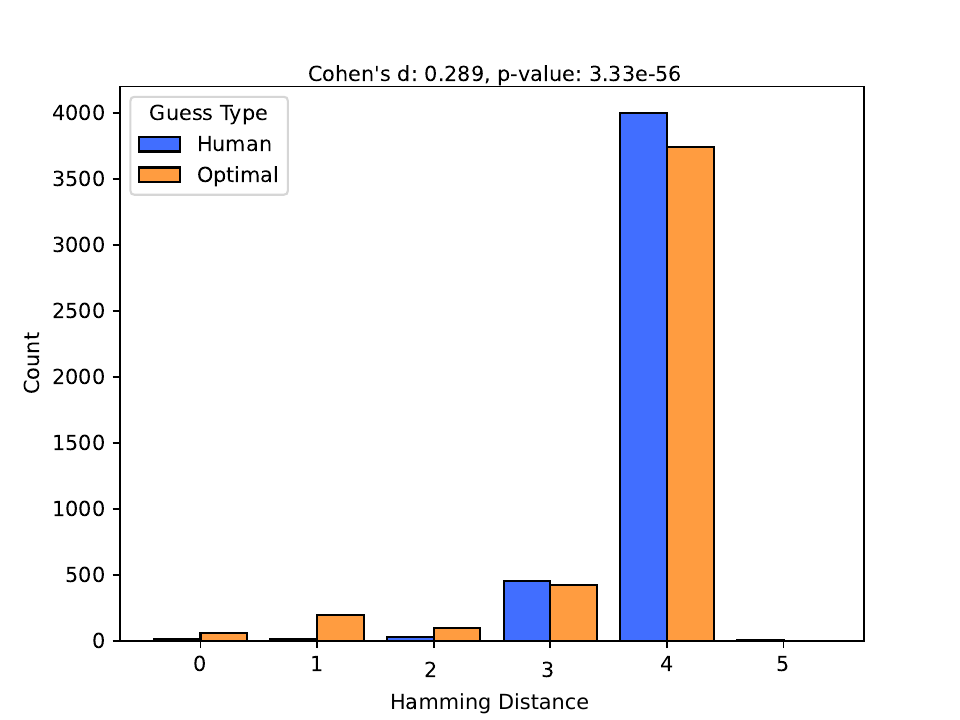} \\
        \parbox{0.45\textwidth}{\small\centering (e) Hamming distance between human guesses and near-optimal guesses for 0g0y5b: humans play very obviously suboptimally by reusing characters they know are not there.} &
        \parbox{0.45\textwidth}{\small\centering (f) Hamming distance between human guesses and near-optimal guesses for 2g2y1b: humans play suboptimally by using new characters.}
    \end{tabular}
    \caption{$(c_g, c_y, c_b)$: the number of ``green" guesses (correct letter in the correct place) is denoted with $c_g$, the number of ``yellow" guesses (correct letter in the incorrect place) is denoted with $c_y$, and the number of letter guesses that are incorrect is $c_b$. \label{results}}
\end{figure*}

In Figure~\ref{results}, we present some observations on the results of our comparison of human play with near-optimal play for specific configurations. We observe that in many, though not all, cases, humans are biased towards their previous guesses. This is particularly pronounced when there is a lot of free choice. This indicates that human gameplay, which combines creativity with optimization, shows more evidence of lack of creativity when there is more free choice. Figures~\ref{box1} and~\ref{box2} in the Appendix display the overall patterns: for more constrained positions, differences between near-optimal moves and actual moves tend to be smaller.

We additionally report that the optimal guess rhymes with the previous guess 7.3\% of the time, but humans make a guess that rhymes with the previous guess 9.3\% of the time (p-value $< 0.001$).
\subsection{Relationship between semantic distance and other measures}

For valid Wordle word pairs, the Hamming distance and the Levenshtein distance are strongly related (Pearson-r=0.95 for all pairs, Pearson-r=0.81 for character difference less than 5). 

\begin{figure}
\includegraphics[width=\columnwidth]{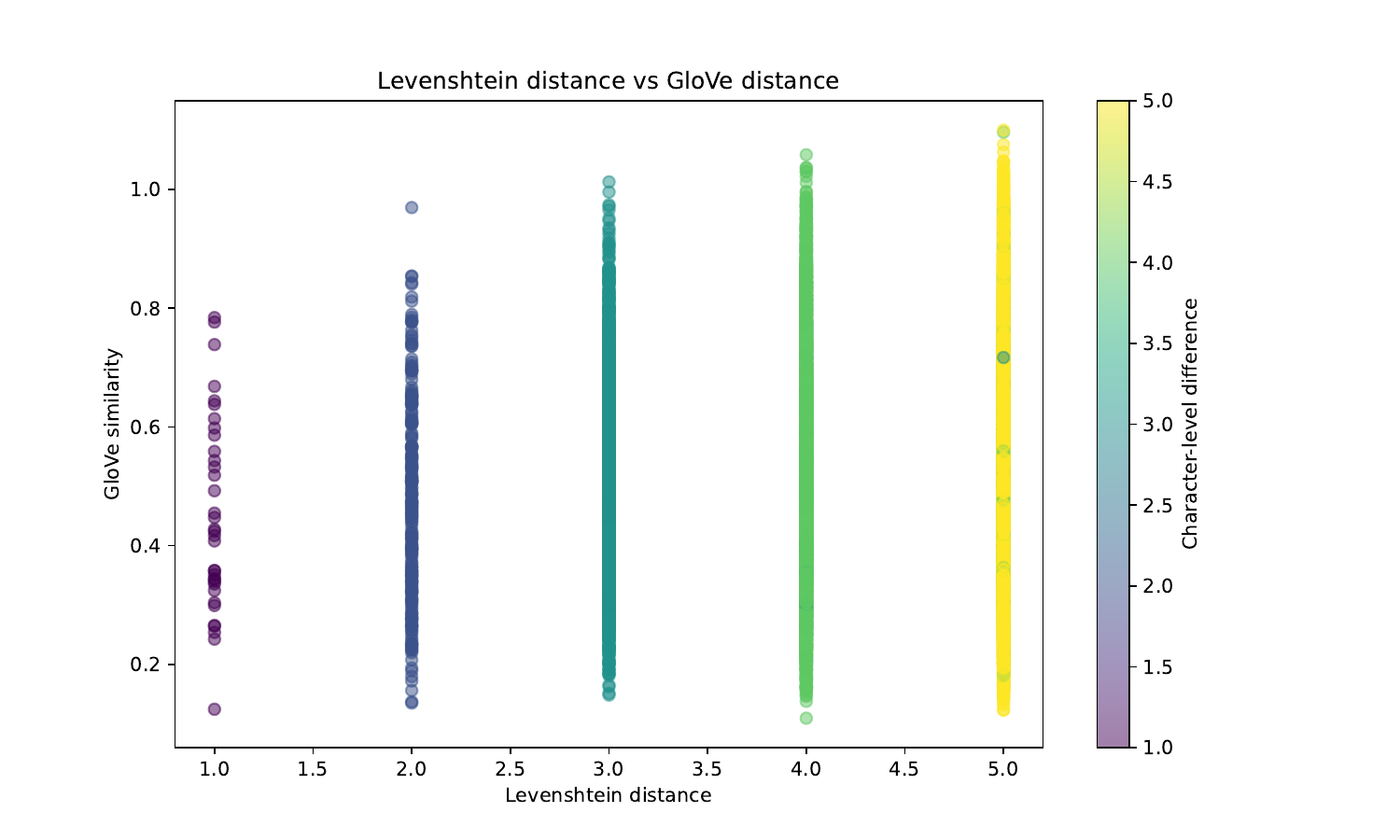}
\caption{GloVe similarity between guesses vs. Levenshtein Distance}
\label{GloVeLevenstain}
\end{figure}

The relationship between character-based differences and semantic distance ($1-cos(v_a, v_b)$ where $v_a$ and $v_b$ are the GloVe vectors) is weaker: Pearson-r= 0.06 (Figure~\ref{GloVeLevenstain}), suggesting that the effects of semantic distance and orthographic distance likely operate separately.


\section{Conclusions}
Human gameplay in Wordle exhibits a bias toward previous guesses semantically, orthographically, and phonologically (e.g., sharing the last syllable). The bias is systematic, indicating human gameplay does not merely randomly deviate from optimal play.

In Wordle, we demonstrate an environment where phenomena like word association can be studied in a setting that is not as varied as natural speech, but where people are not merely performing a task, as in the word association task (cf.~\citet{hamilton2020revolution}). Environments such as Wordle can be of interest when studying other cognitive patterns.
\bibliography{wordle1}

\section{Limitations}
We rely on publicly available data voluntarily posted by users on Reddit. In principle, games posted by 
Reddit users can differ systematically from data 
sampled randomly from a different population of 
interest. This is due to selection effects as well as 
the fact that the population of Reddit users, and 
specifically the Wordle subreddit, could differ by 
demographics and linguistic background, among 
other things, from other populations of interest.
We limit our study to the game of Wordle in 
English. Our findings may not generalize to other 
games in other languages.

\clearpage
\appendix

\clearpage
\newpage
\begin{figure*}
    \centering
    \begin{tabular}{@{}cc@{}}
        \includegraphics[width=0.45\textwidth]{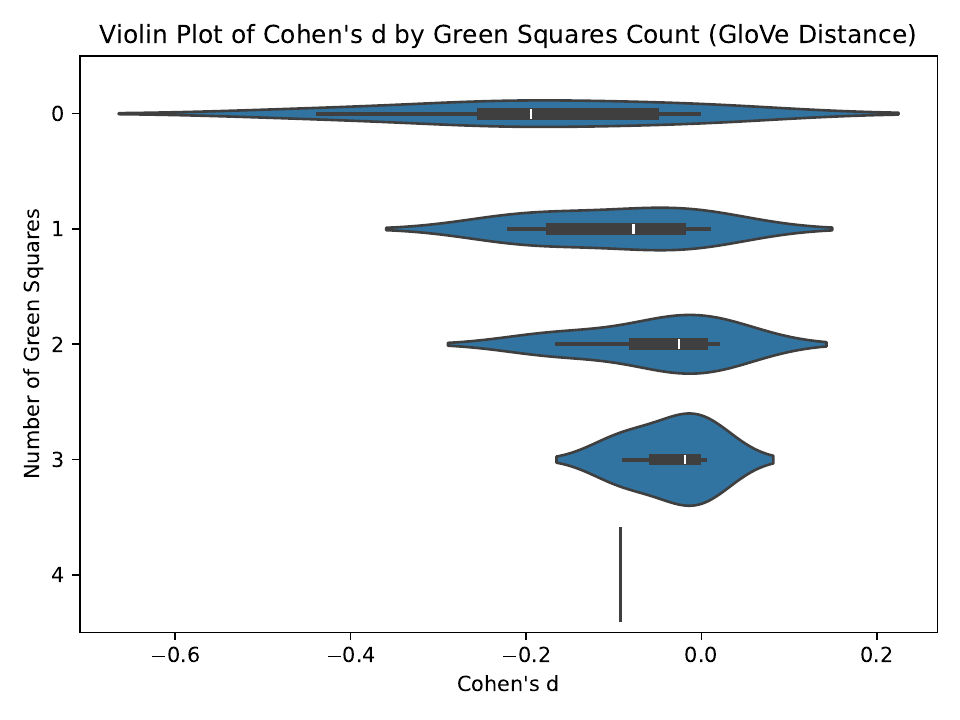} &
        \includegraphics[width=0.45\textwidth]{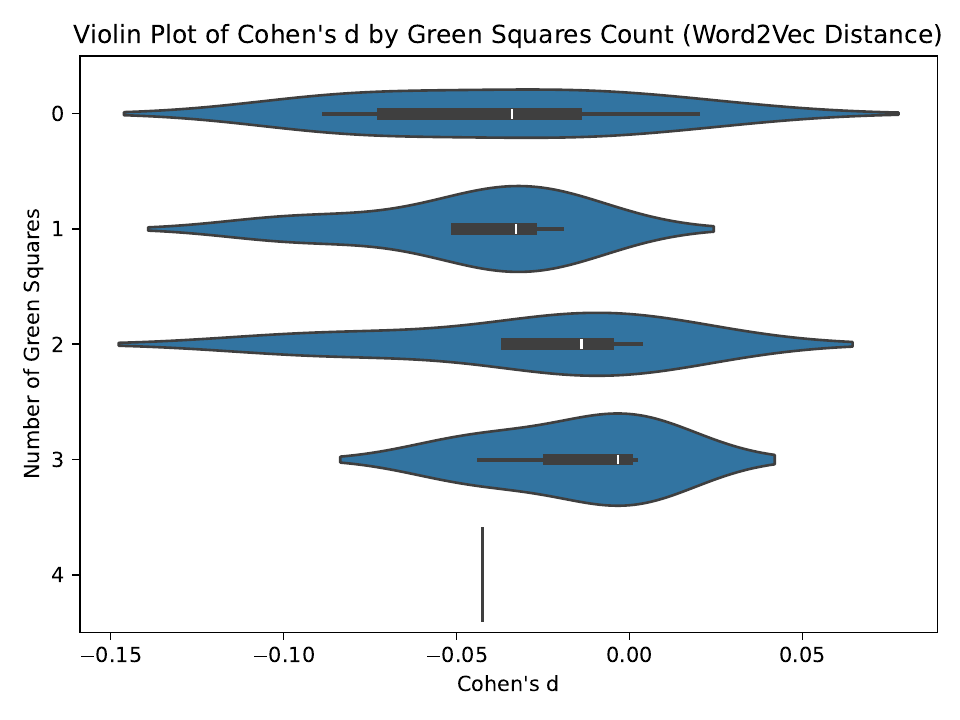} \\
        \parbox{0.45\textwidth}{\small\centering (a) Violin plot of Cohen $d$ by \emph{green} squares (GloVe).} &
        \parbox{0.45\textwidth}{\small\centering (b) Violin plot of Cohen $d$ by \emph{green} squares (Word2Vec).} \\[1cm]
        
        \includegraphics[width=0.45\textwidth]{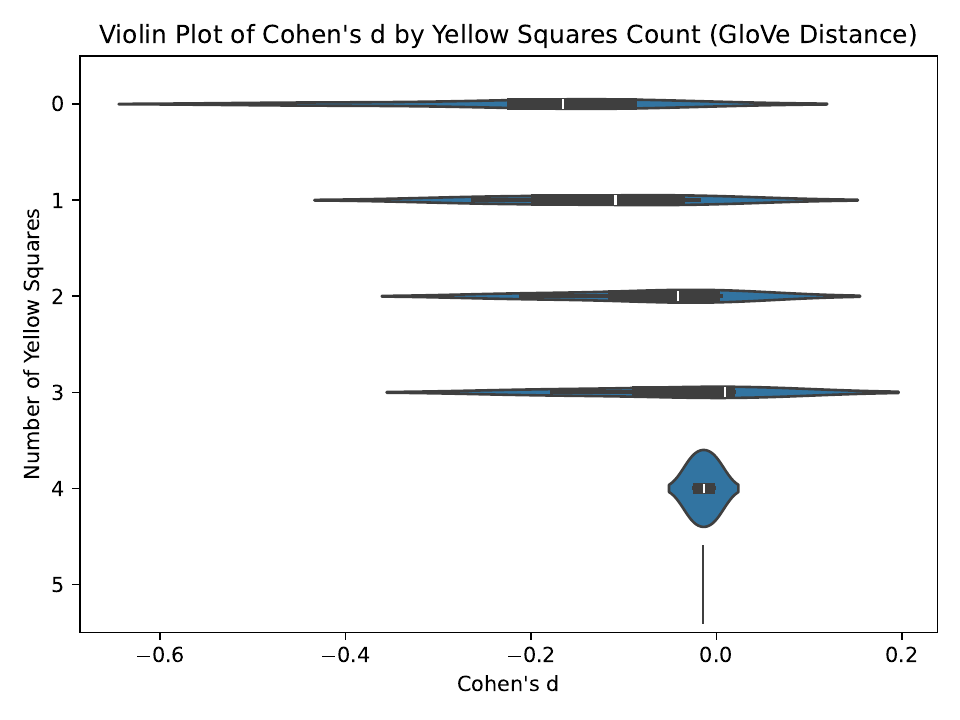} &
        \includegraphics[width=0.45\textwidth]{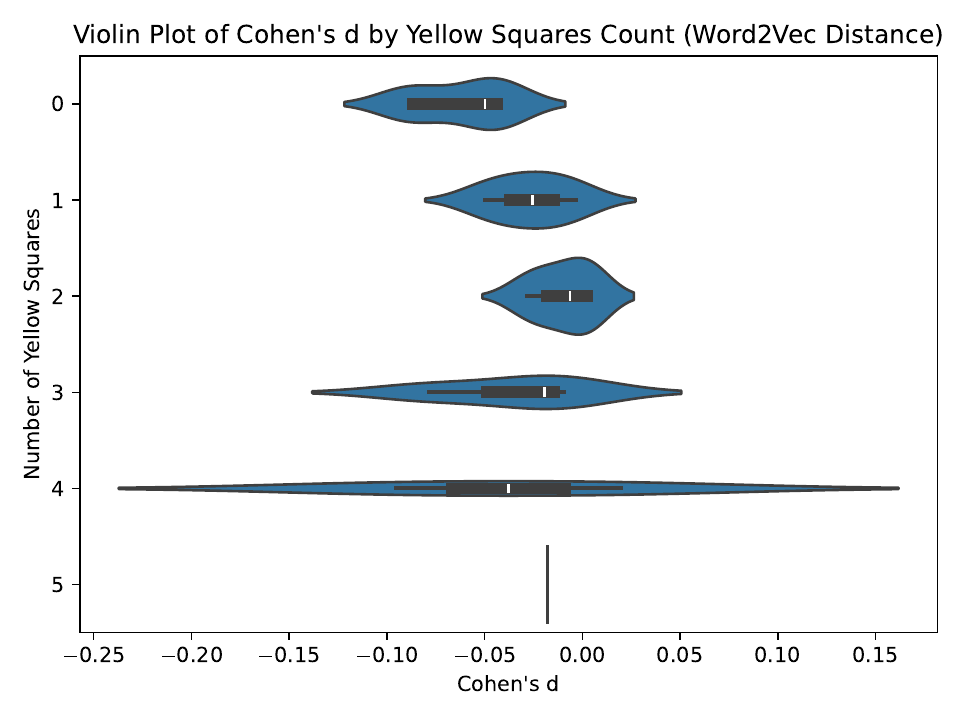} \\
        \parbox{0.45\textwidth}{\small\centering (c) Violin plot of Cohen $d$ by \emph{yellow} squares (GloVe).} &
        \parbox{0.45\textwidth}{\small\centering (d) Violin plot of Cohen $d$ by \emph{yellow} squares (Word2Vec).} \\[1cm]
        
        \includegraphics[width=0.45\textwidth]{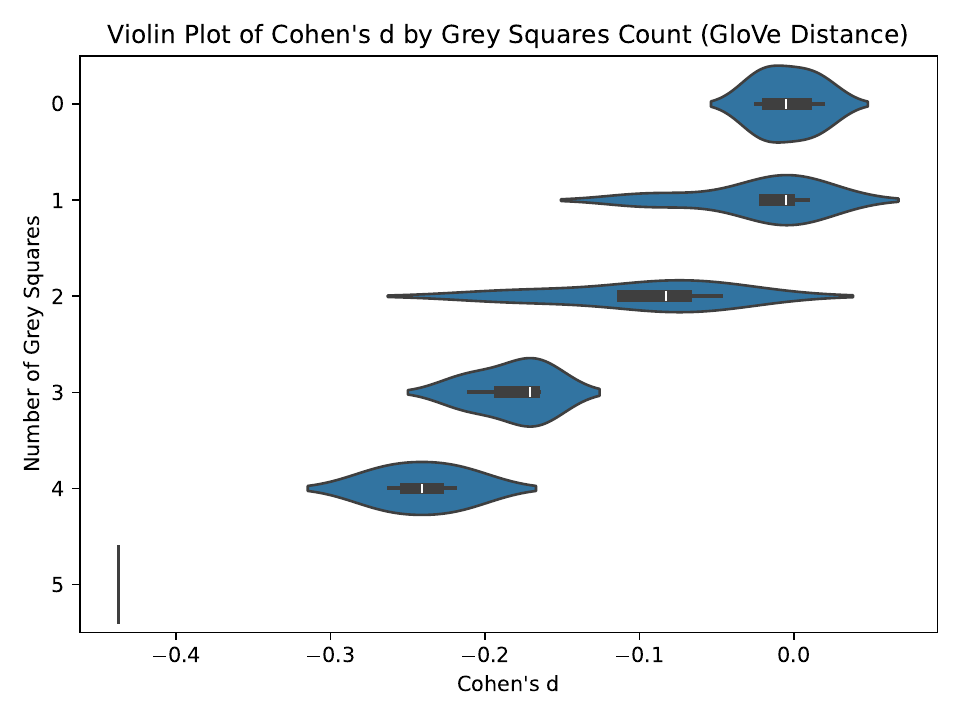} &
        \includegraphics[width=0.45\textwidth]{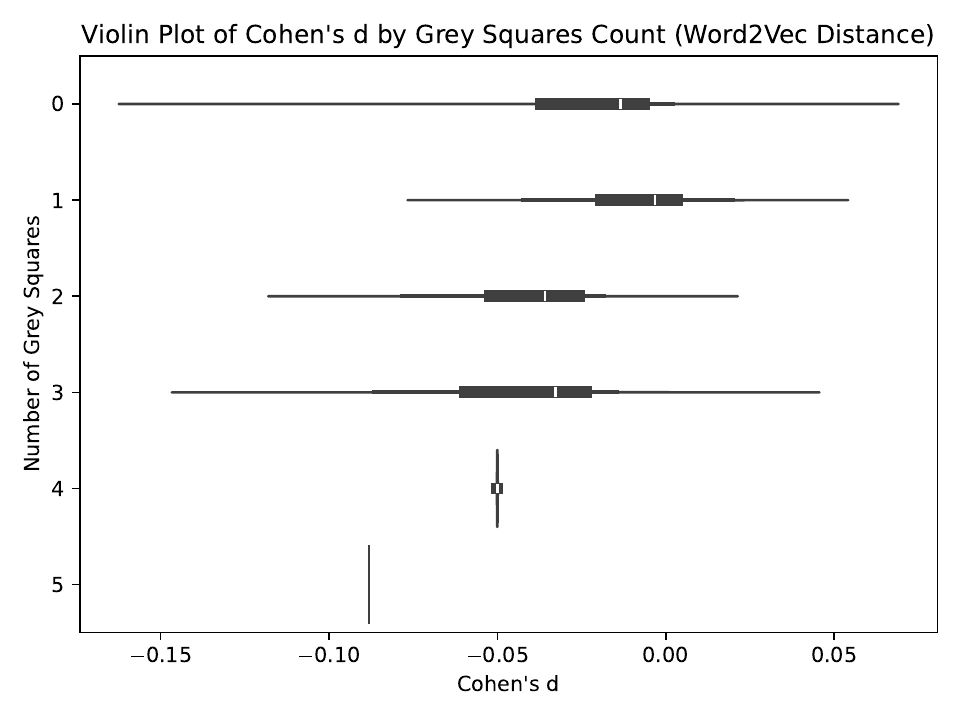} \\
        \parbox{0.45\textwidth}{\small\centering (e) Violin plot of Cohen $d$ by \emph{grey} squares (GloVe).} &
        \parbox{0.45\textwidth}{\small\centering (f) Violin plot of Cohen $d$ by \emph{grey} squares (Word2Vec).} \\[1cm]
        
    \end{tabular}
    \caption{Distances between neighboring guesses, by amount of constraint on guess}
    \label{box1}
\end{figure*}

\newpage

\clearpage
\newpage

\begin{figure*}
    \centering
    \begin{tabular}{@{}cc@{}}

        \includegraphics[width=0.45\textwidth]{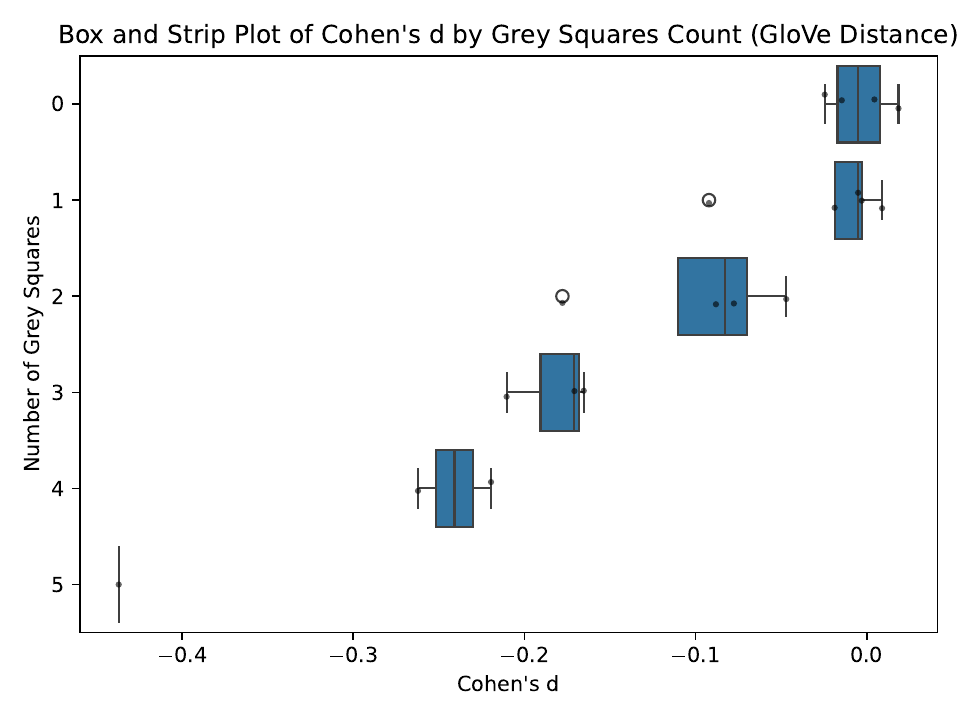} &
        \includegraphics[width=0.45\textwidth]{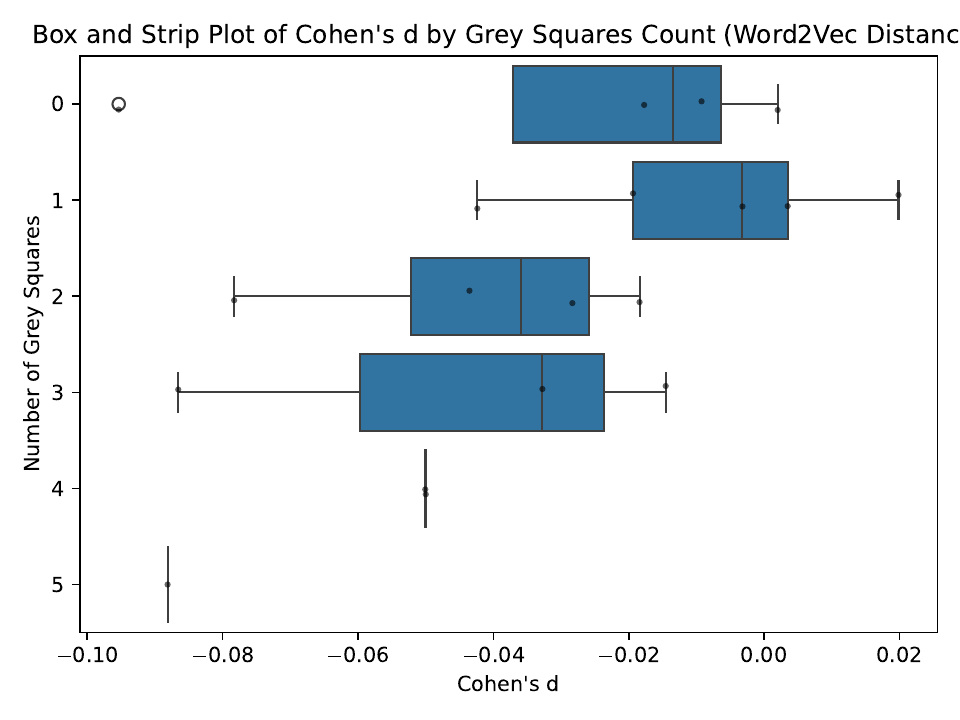} \\
        \parbox{0.45\textwidth}{\small\centering (a) Box \& strip plot of Cohen $d$ by \emph{grey} squares (GloVe).} &
        \parbox{0.45\textwidth}{\small\centering (b) Box \& strip plot of Cohen $d$ by \emph{grey} squares (Word2Vec).}
    \end{tabular}
    \caption{Distances between neighboring guesses, by amount of constraint on guess}
\label{box2}
\end{figure*}

\end{document}